\documentclass[journal]{IEEEtran}

\usepackage[pagebackref=true,breaklinks=true,letterpaper=true,colorlinks,bookmarks=false]{hyperref}
\usepackage{times}
\usepackage{epsfig}
\usepackage{graphicx}
\usepackage{subcaption}
\usepackage{amsmath}
\usepackage{amssymb}
\usepackage{multicol}
\usepackage{bm}
\usepackage{adjustbox}
\usepackage{booktabs}
\usepackage{epsfig}
\usepackage{tabularx}
\usepackage{tablefootnote}
\usepackage{color}
\usepackage{pifont}
\usepackage[accsupp]{axessibility}  % Improves PDF readability for those with disabilities.
\newcommand{\xmark}{\ding{55}}
\usepackage{multirow}
\usepackage{pgfplots}
\usepackage{tikz}
\usepackage{hyperref}
\usepackage{orcidlink}
\ifCLASSINFOpdf
\else
\fi

\begin{document}
\title{ELGC-Net: Efficient Local-Global Context Aggregation for Remote Sensing Change Detection}

\author{Mubashir~Noman$^1$~
        Mustansar~Fiaz$^2$~
        Hisham~Cholakkal$^1$~
        Salman~Khan$^{1,3}$~
        and~Fahad~Shahbaz~Khan$^{1,4}$
        \vspace{0.5em}
        \\
        % <-this % stops a space
$^1$Mohamed bin Zayed University of AI \quad $^2$IBM Research \quad $^3$Australian National University \\$^4$Linköping University
\vspace{-1.0em}
}
\maketitle

\begin{abstract}
Deep learning has shown remarkable success in remote sensing change detection (CD), aiming to identify semantic change regions between co-registered satellite image pairs acquired at distinct time stamps. However, existing convolutional neural network (CNN) and transformer-based frameworks often struggle to  accurately segment semantic change regions.  Moreover, transformers-based methods with standard self-attention suffer from quadratic computational complexity with respect to the image resolution, making them less practical for CD tasks with limited training data. To address these issues, we propose an efficient change detection framework, ELGC-Net, which leverages rich contextual information to precisely estimate change regions while reducing the model size. Our ELGC-Net comprises a Siamese encoder, fusion modules, and a decoder. The focus of our design is the introduction of an Efficient Local-Global Context Aggregator (ELGCA) module within the encoder, capturing enhanced global context and local spatial  information through a novel pooled-transpose (PT) attention and depthwise convolution, respectively. The PT attention employs  pooling operations for robust feature extraction and minimizes computational cost with transposed attention.  %Furthermore, we fuse contextual information obtained from different channel subsets, further easing computational complexity. 
Extensive experiments on three challenging CD datasets demonstrate that ELGC-Net outperforms existing methods.
%Compared to the best transformer-based CD approach, ELGC-Net achieves a 9.55\% and 1.35\% gain in intersection over union (IoU) metric on DSIFN-CD and LEVIR-CD datasets, respectively, while significantly reducing trainable parameters.
Compared to the recent transformer-based CD approach (ChangeFormer), ELGC-Net achieves a 1.4\% gain in intersection over union (IoU) metric on the LEVIR-CD dataset, while significantly reducing trainable parameters.
Our proposed ELGC-Net sets a new state-of-the-art performance in remote sensing change detection benchmarks.  Finally, we also introduce ELGC-Net-LW, a  lighter variant with significantly reduced computational complexity,   suitable for resource-constrained settings, while achieving comparable performance. %Our source code and trained models will be made publicly available. 
Our source code is publicly available at \url{https://github.com/techmn/elgcnet}.

\end{abstract}


%\def\thefootnote{*}\footnotetext{Fahad Khan is the corresponding author.}

% \fi
\IEEEpeerreviewmaketitle

\section{Introduction}
\IEEEPARstart{R}{emote}  sensing change detection is a challenging task where the objective is to detect semantic changes between satellite image pairs taken at different time stamps. It finds diverse applications such as environmental monitoring, urban planning \cite{yin2021integrating}, land cover mapping, and disaster assessment \cite{kucharczyk2021remote}. With the increasing availability of high-resolution satellite imagery, the demand for efficient and accurate change detection methods to process this huge amount of data has grown significantly in recent years.

\begin{figure*}[t]
\begin{center}
{\includegraphics[width=1\linewidth, keepaspectratio] {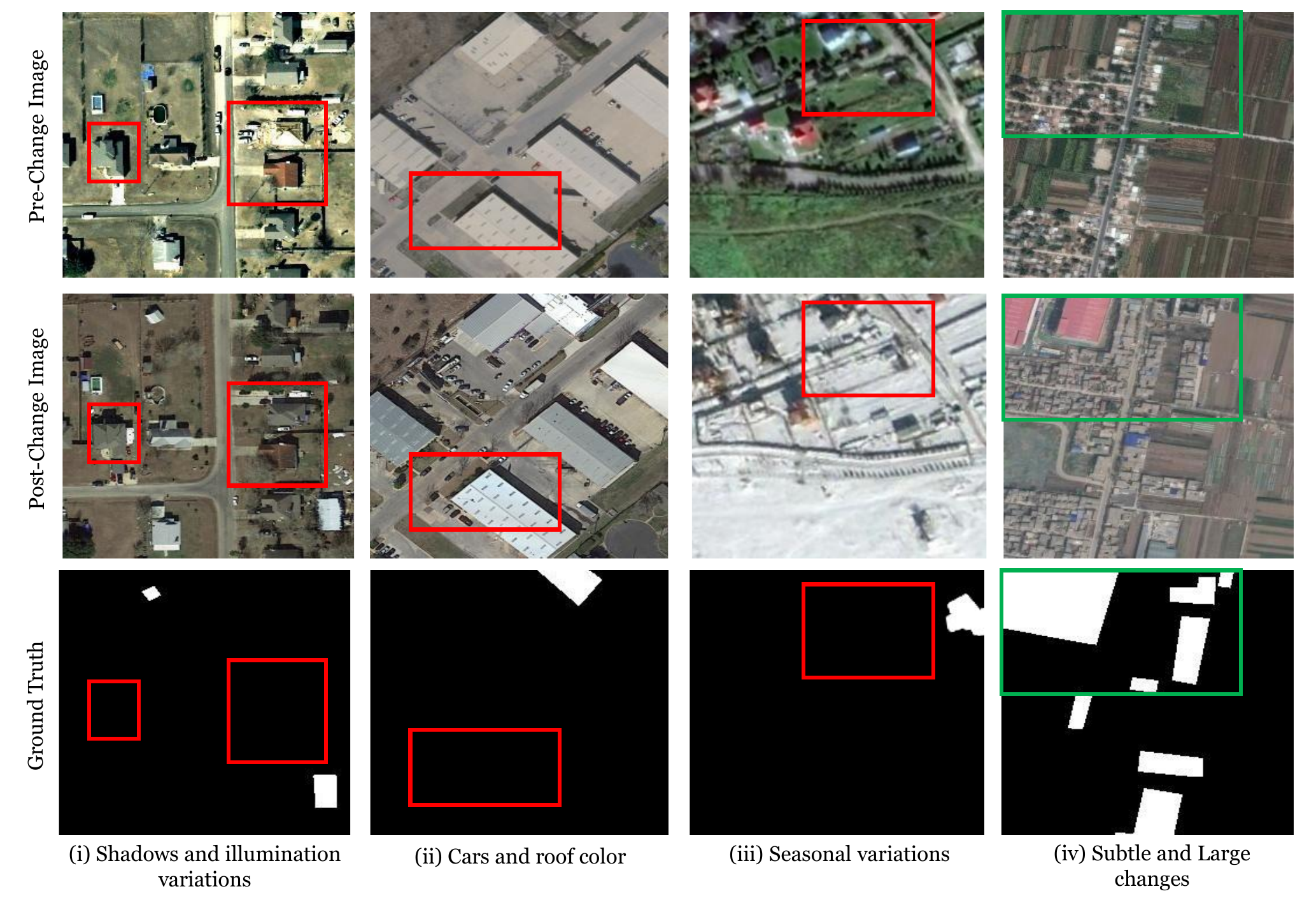}}
\end{center}
\vspace{-0.5em}
\caption{Example image pairs depicting various challenges related to remote sensing change detection and the corresponding ground-truth segmentation masks. It is highly challenging to ignore semantically irrelevant changes (shown in red boxes) such as (i) shadows and illumination variations, (ii) movable cars and building roof changes (iii) seasonal variations, while accurately segmenting   (iv) subtle and large semantic changes (shown in green box).
%such as irregular building shapes, .  
%The leftmost column shows the challenges related to shadows and illumination variations. In the second from the left column, the cars and color of the building roof are not considered as a semantic change. The second from the right column depicts the seasonal variations challenge, and the rightmost column shows the trees,  irregular building shapes, and a lot of subtle change regions.
}
%\vspace{-0.2em}
\label{fig_challenges}
\end{figure*}

Remote sensing change detection faces several challenges, such as environmental variations, complex and irregular geometrical structures, variations in object scales \cite{li2022transunetcd}, and the high-resolution nature of satellite imagery. These challenges introduce various complexities when accurately identifying changes between image pairs. For instance, Fig.~\ref{fig_challenges} (columns 1-3) illustrates semantically irrelevant changes that need to be ignored by the change detection (CD) algorithm (indicated by red boxes), such as (i) shadows and illumination variations, (ii) displacement of movable objects like cars and roof color changes, and (iii) seasonal variations in land coverings, while (iv) accurately segmenting both subtle and major semantic changes between image pairs (indicated by green boxes in the last column).  
Thus, it is desired to develop an accurate CD method that can effectively learn the semantic information and accurately detect the change regions while suppressing irrelevant ones.
 {Traditional approaches utilize feature difference technique and obtain binary mask via thresholding  \cite{rosin1998thresholding, rosin2003evaluation}. However, such approaches are unable to capture the underlying characteristics of the objects and fail to discriminate between the semantic and noisy changes. Numerous methods have been proposed that are based on hand-crafted features and utilize different classifiers including decision trees \cite{im2005change},   change vector analysis \cite{bovolo2011framework}, support vector machine \cite{volpi2013supervised}, and clustering approaches \cite{aiazzi2013nonparametric, shang2014change} for better discrimination.
Nevertheless, these classical approaches exhibit various limitations and are incapable of capturing rich feature representations.}

%##################################

On the other hand,  convolutional neural network (CNN) based CD methods \cite{daudt2018_fcsiam, zhang2018triplet, chen2020dasnet, zhang2020deeply, chen2020spatial} utilize Siamese architecture to address the CD problem. These approaches typically exploit architectural components such as dilated convolutions, multi-scale features, and attention mechanisms in their framework. However, these CNN-based approaches often struggle to capture global contextual representations, which is likely to deteriorate the CD performance. 
 {
In contrast, transformer-based approaches \cite{changeformer, yan2022fully}
employing standard self-attention \cite{vaswani2017attention} often struggles to effectively incorporate local contextual information for accurate segmentation of subtle changes at irregular boundaries.
Furthermore, while state-of-the-art transformer-based change detection frameworks \cite{changeformer} have achieved impressive results, they come with inherent challenges. These approaches generally rely on standard self-attention, which exhibits quadratic complexity with the number of tokens in the encoder. Consequently, the requirement for a large number of parameters, extensive memory usage, and high FLOPs (floating-point operations per second) make these methods less suitable for practical remote sensing change detection applications.
%In contrast, transformer-based approaches \cite{changeformer, yan2022fully} capture global contextual information  by utilizing self-attention \cite{vaswani2017attention}  which may restrain their learning capabilities to encode local dependencies for accurate segmentation of subtle changes at irregular boundaries. %This is mainly due to dominant global contextual information leading toward the suboptimal CD solution.
Hence, it is desired to capture both local and global contextual information to effectively detect subtle as well as significant structural changes between image pairs.  

}

 {
In order to handle the above mentioned challenges, we propose a Siamese-based efficient change detection framework named  ELGC-Net to accurately identify semantic changes in satellite image pairs taken at different time stamps.
The focus of our design is to the introduction of a novel efficient local-global context aggregator (ELGCA) module that can capture both local and global contextual information while addressing the limitations of standard self-attention.
Specifically, our ELGCA strives to (i) capture global contextual information through pooled-transpose (PT) attention, (ii) encode local spatial context using depth-wise convolution, and (iii) multi-channel feature aggregation.  The PT attention leverages pooling operations for robust feature extraction and reduces the computational cost with transposed attention. 
We extensively evaluate ELGC-Net on three diverse change detection benchmark datasets, validating its superiority over existing methods in literature.
%transformer-based methods such as ChangeFormer \cite{changeformer} and TransUNetCD \cite{li2022transunetcd}. 
Our method achieves state-of-the-art performance while significantly reducing the model size, as demonstrated by the comprehensive quantitative and qualitative analysis. 
In addition to the  ELGC-Net, we introduce its lightweight variant called ELGC-Net-LW that requires only 6.78M parameters while achieving comparable performance to recent ChangeFormer \cite{changeformer}, which utilizes 41M parameters (as shown in Fig. 2 in the manuscript). ELGC-Net-LW serves as a resource-efficient alternative, catering to scenarios with limited computational resources without compromising on CD accuracy. The key contributions of our method are summarized as:
}

\begin{itemize}
\item  {We propose a Siamese-based efficient change detection framework named as ELGC-Net to accurately identify semantic changes between the bi-temporal satellite images.}
\item  {Instead of modeling dense dependencies, our ELGCA module proposes to capture both local and global relationships through depthwise convolution and pooled-transpose attention, respectively, while reducing computational complexity.}
\item  {Our extensive experiments on three CD datasets demonstrate the robustness and efficiency of our approach while achieving state-of-the-art performance. Moreover, our lightweight variant called ELGC-Net-LW obtains comparable performance compared to recent ChangeFormer \cite{changeformer} requiring only 6.78M parameters.}    
\end{itemize}

The remainder of this paper is structured as follows. Section II presents the related works and Section III discusses about the baseline framework. Section IV presents an in-depth discussion of the proposed method and our Efficient Local-Global Context Aggregator (ELGCA) module. Section V presents the experimental setup, results, and performance comparison with existing methods, and finally, Section VI concludes the paper.

%##################################

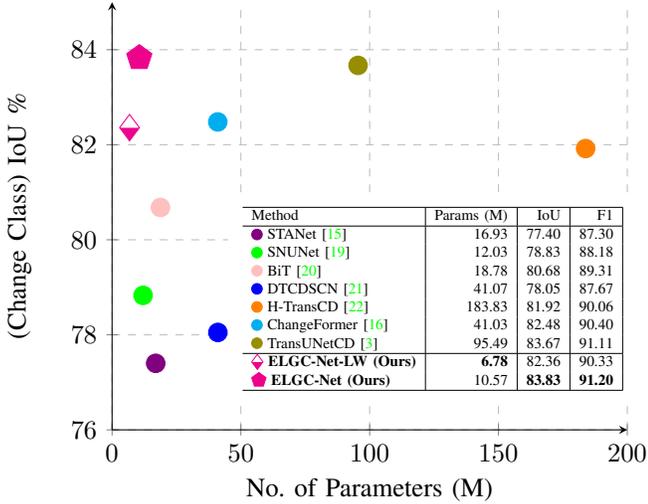
\begin{figure}[t!]
\centering
\resizebox{\linewidth}{!}{%
\begin{tikzpicture}
\begin{axis}[
axis lines = left,
ymin=76, ymax=85,
xmin=0, xmax=200,
xlabel= No. of Parameters (M),
ylabel = (Change Class) IoU \%,
xmajorgrids=true,
ymajorgrids=true,
grid style=dashed,
]
\coordinate (legend) at (axis description cs:1.02,0.07);
%%%% Add Methods name here %% original line axis description cs:0.99,0.006
%% STANet
\addplot[only marks,
mark=otimes*, violet,
mark size=3.5pt
]
coordinates {
(16.93,77.4)};\label{plot:STANet}
%SNUNet
\addplot[only marks,
mark=otimes*, green,
mark size=3.5pt
]
coordinates {
(12.03,78.83)};\label{plot:SNUNet}
%% BiT
\addplot[only marks,
mark=otimes*, pink,
mark size=3.5pt
]
coordinates {
(18.78,80.68)};\label{plot:BiT}
%% DTCDSCN
\addplot[only marks,
mark=otimes*, blue,
mark size=3.5pt
]
coordinates {
(41.07,78.05)};\label{plot:DTCDSCN}
%% ChangeFormer
\addplot[only marks,
mark=otimes*, cyan,
mark size=3.5pt
]
coordinates {
(41.03,82.48)};\label{plot:ChangeFormer}
%% Hybrid-TransCD
\addplot[only marks,
mark=otimes*, orange,
mark size=3.5pt
]
coordinates {
(183.83,81.92)};\label{plot:Hybrid-TransCD}
%% TransUNetCD
\addplot[only marks,
mark=otimes*, olive,
mark size=3.5pt
]
coordinates {
(95.49,83.67)};\label{plot:TransUNetCD}
%% ELGCNet-LW (Ours)
\addplot[only marks,
mark=halfdiamond*, magenta,
mark size=5pt
]
coordinates {
(6.78, 82.36)};\label{plot:ELGCNet-LW}
%% ELGCNet (Ours)
\addplot[only marks,
mark=pentagon*, magenta,
mark size=5pt
]
coordinates {
(10.57, 83.83)};\label{plot:ELGCNet}
\end{axis} % south east
\node[draw=none,fill=white, anchor= south east] at
(legend){\resizebox{5.2cm}{!}{
\begin{tabular}{l|r|r|r|r}
\hline
Method & Params (M) & IoU & F1 \\ \hline
\ref{plot:STANet} STANet~\cite{chen2020spatial} & 16.93 & 77.40 & 87.30 \\
\ref{plot:SNUNet} SNUNet~\cite{fang_snunet} & 12.03 & 78.83 & 88.18 \\
\ref{plot:BiT} BiT~\cite{chen2021_bit} & 18.78 & 80.68 & 89.31 \\
\ref{plot:DTCDSCN} DTCDSCN~\cite{liu2020building} & 41.07 & 78.05 & 87.67 \\
\ref{plot:Hybrid-TransCD} H-TransCD~\cite{ke2022hybrid} & 183.83 & 81.92 & 90.06  \\
\ref{plot:ChangeFormer} ChangeFormer~\cite{changeformer} & 41.03 & 82.48 & 90.40 \\
\ref{plot:TransUNetCD} TransUNetCD~\cite{li2022transunetcd} & 95.49 & 83.67 & 91.11 \\
\hline
\ref{plot:ELGCNet-LW} \textbf{ELGC-Net-LW (Ours)} & \textbf{6.78} & 82.36 & 90.33 \\
\ref{plot:ELGCNet} \textbf{ELGC-Net (Ours)} & 10.57 & \textbf{83.83} & \textbf{91.20} \\
\hline
\end{tabular} }};
\end{tikzpicture}
}\vspace{-0.3cm}
\caption{Accuracy (IoU) vs. model size (params) comparison with existing methods on  on LEVIR-CD.  Our ELGCNet achieves state-of-the-art performance
%while requiring only 9\% of model parameters compared to the existing state-of-the-art TransUNetCD \cite{li2022transunetcd}.
while having $9\times$ less number of model parameters compared to the existing state-of-the-art TransUNetCD \cite{li2022transunetcd}.
} \vspace{-0.2cm}
\label{fig:params_vs_iou_latex}
\end{figure}
%%---------------------
%##################################

%##################################
\begin{figure*}[!t]
\begin{center}
{\includegraphics[width=1\linewidth, keepaspectratio] {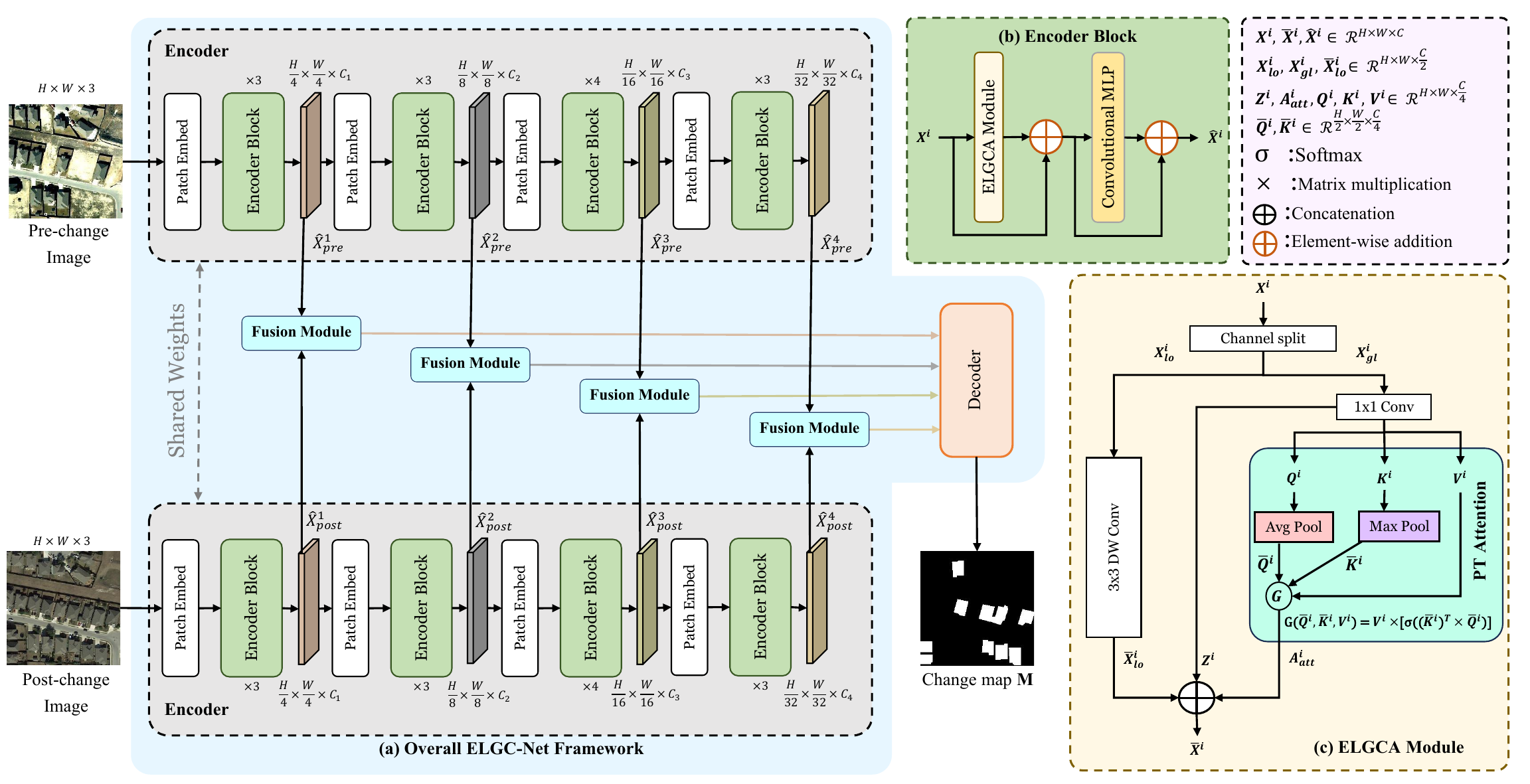}}
\end{center}
%\caption{Overall architecture of the proposed network. The pre- and post-change images are input to the shared encoder to extract features using encoder blocks at four scale levels. At each scale level, features are passed from a linear layer and fused via a convolution layer. These fused feature maps at each scale are merged in the light weight decoder and passed from few convolution and transposed convolution layers to obtain up sampled features. Finally, prediction layer of decoder detects the change regions in the up sampled feature maps.}
\caption{Overall architecture of the proposed CD framework. (a) The complete network architecture is presented, illustrating the  pre- and post-change input images to the shared encoder. The encoder blocks extract features at four stages and these features are merged by a simple fusion module comprising linear projection, feature concatenation, and a $1\times1$ convolution.
These fused feature maps at each stage are then merged in the decoder, which includes several convolutions and transposed convolution layers to obtain upsampled features. The upsampled feature maps are used in the prediction layer to obtain the final change map. (b) The structure of our Encoder block is shown, featuring the Efficient Local-Global Context Aggregation (ELGCA) module and a convolutional MLP. (c) A detailed view of the proposed ELGCA module performing the following key operations: (i) capturing local spatial context  using a 3x3 depth-wise convolution ($\bar X^i_{lo}$)
and (ii) global context aggregation ($A^i_{att}$) through a pooled-transpose (PT) attention operation,
and multi-channel feature aggregation using a 1x1 convolution ($Z^i$). 
To enhance the efficiency of our PT attention, 
we perform transposed attention ($G$) having linear complexity with the number of tokens, on pooled $Q^i$ and $K^i$ tokens (denoted as $\bar{Q}^i$, $\bar{K}^i$) on a sub-set of channels ($C/4$). 
The aforementioned operations within our ELGCA module are performed in parallel on different groups (subsets) of channels obtained through channel splitting, leading to improved computational efficiency.}
\label{fig_overall_architecture}
\end{figure*}
%##################################

\section{Related Work}

 {

As mentioned earlier, the automatic CD has been widely studied over the past decade and various CD methods have been proposed \cite{coppin2004digital, singh1989review, radke2005image}. Earlier, conventional methods strive to identify the changes via computing the image difference techniques \cite{rosin1998thresholding, rosin2003evaluation}, in the bi-temporal images. To do so, these approaches typically generate a change map by identifying the pixels that have substantial differences and then employ a threshold to finally obtain the segmentation mask. Nevertheless, these approaches exhibit few benefits due to the incapabilities of the RGB features to discriminate between semantic and irrelevant changes. In addition, different classification approaches have been proposed which employ hand-crafted features \cite{radke2005image}  and discriminate the change features from the temporal images via considering statistical modeling \cite{bazi2005unsupervised, zanetti2015rayleigh}, decision tree \cite{im2005change},   change vector analysis \cite{bovolo2011framework}, support vector machine \cite{volpi2013supervised}, and clustering approaches \cite{aiazzi2013nonparametric, shang2014change}.
}

 {
Later, deep learning approaches based on deep Convolutional Neural Networks (CNNs)   have shown significant performance improvement on numerous computer vision task such as segmentation \cite{long2015fully, fiaz2020video}, object tracking \cite{fiaz2020improving}, action recognition \cite{feichtenhofer2017spatiotemporal} and many more. 
%They have also shown great progress in  Land cover change detection (LCCD) \cite{lv2023multi, lv2023hierarchical, lv2023spatial}.
The CNNs have shown their capability to better capture the underlying characteristics and are the key factor in utilizing them for the remote sensing change detection field \cite{shi2020cd_review}. 
Various deep learning methods have also been explored in the field of remote sensing change detection \cite{zhao2018building, liu2019instance, ham2018semantic, khan2017learning, alcantarilla2018street, daudt2018_fcsiam, papadomanolaki2019detecting}.
Ham et al. \cite{ham2018semantic}  used a deconvolutional network as a semantic segmentation algorithm to detect unregistered buildings from UAVs (Unmanned Aerial Vehicles) images.
Zhao et al. \cite{zhao2018building} and Liu et al. \cite{liu2019instance} employed Mask-RCNN method for semantic segmentation of buildings and outdoor sports venues, respectively.  BiDateNet \cite{papadomanolaki2019detecting} proposed a  deep learning framework based on U-Net architecture to obtain a change map.
Guo et al. \cite{guo2018learning} configured a Fully Convolutional Network in the form of a Siamese Network to detect changes in continuous images. Khan et al. \cite{khan2017learning} used a CNN-based weakly supervised technique to directly detect and localize the change detection for the input image pairs.  CDNet \cite{alcantarilla2018street}, FC-EF, FC-Siam-diff, and FC-Siam-Conc \cite{daudt2018_fcsiam} employed boundary decisions to obtain the change map.
}

 {Although CNN methods exhibit better performance in the field, they strive to capture the long-range dependencies whereas transformers may struggle in capturing accurate localization due to lack of fine details \cite{li2022transunetcd}.}
This motivated the researchers to combine the large receptive fields of transformers with the local contextual features of convolutions to obtain discriminative feature representations \cite{noman2023scratchformer, rs_survey_aman_2023}.
Jiang et al. \cite{jiang2020pga} introduce a co-attention module to enhance the features for better detection of change regions. The authors utilize the Siamese network in encoder-decoder architecture and use the VGG-16 network as a backbone feature extractor. Afterward, extracted features are enhanced by utilizing the co-attention module and input to the decoder for change map prediction.
Lie et al. \cite{li2022transunetcd} propose a hybrid transformer model that uses a Siamese CNN encoder to extract features and concatenate the encoded feature of two streams. Then, they capture the global contextual information using a transformer and use a cascading up-sampling decoder to predict the change map.
Chen et al. \cite{chen2021_bit} propose a bi-temporal image transformer (BIT) network to model the long-range context information for CD task. The BIT utilizes the ResNet18 \cite{kaiming2016_resnet} to extract features from bi-temporal images followed by a transformer encoder to capture global context relationships in bi-temporal images. A decoder then projects back the encoded features to spatial space and obtains a binary change map.

Song et al. \cite{song2022mstdsnet} introduce a hybrid model based on ResNet18 \cite{kaiming2016_resnet} backbone feature extractor and multiscale SwinTransformer \cite{liu2021Swin} for enhancing the extracted feature maps.
Ke et al. \cite{ke2022hybrid} propose a hybrid transformer module to leverage the Siamese-based CNN backbone. Extracted features from the CNN backbone are enhanced using a hybrid transformer and a cascade decoder predicts the change map.
Bandara et al. \cite{changeformer} introduce a transformer-based encoder to capture better feature representation at multiple scales. The decoder fuses the multiscale features from the bi-temporal image and predicts the binary change mask.
Chen et al. \cite{chen2020dasnet} introduce a dual attention mechanism to capture the discriminative features for better detection of CD regions.
Fang et al. \cite{fang_snunet} propose an ensemble channel attention module (ECAM) to aggregate and refine features at multiple levels to get better CD results.
Zhang et al. \cite{zhang2020deeply} propose a deeply supervised difference discrimination network that takes the extracted bi-temporal features and utilizes a difference discrimination network to predict a change map.
Chen et al. \cite{chen2020spatial} propose a basic and pyramid spatio-temporal attention module to obtain better discriminative features for CD task.

Contrary to the above, we propose an efficient global and local context aggregator that combines the features from  different context aggregators in a parallel manner to obtain rich feature representations while maintaining fewer parameters at multi-scale levels that can discriminate the complex change regions with better accuracy.
%\IEEEpubidadjcol

\section{Base Framework}
Our base framework is a transformer-based Siamese network consisting of an encoder, fusion modules, and a   decoder. The encoder comprises four stages, each incorporating multi-head self-attention layers followed by a convolutional MLP (multi-layer perceptron) network. This base framework takes pre- and post-change images as input and extracts corresponding feature pairs at different stages using a shared encoder.

At each stage, the extracted features undergo a linear layer transformation before being fed to a fusion module. The fusion module encodes the semantic changes between the image pairs by concatenating the features, denoted as $\hat X^i_{pre}$ and $\hat X^i_{post}$, across the channel dimension. Subsequently, a convolutional layer is employed to reduce the number of channels, followed by an activation layer.

The fused features from all stages are then passed to a decoder, where these features are merged and further processed through several convolutions and transposed convolution layers to obtain higher-resolution feature maps. Finally, these upsampled feature maps are utilized in a prediction layer to obtain the final change map.

\section{Method} 
This section describes the overall architecture of the network and explains the ELGCA module in detail.

\subsection{Overall Architecture}
The overall network architecture of the proposed change detection (CD) framework, named as ELGC-Net, is illustrated in Fig.~\ref{fig_overall_architecture}-(a). Similar to the baseline framework, the proposed CD framework consists of a Siamese encoder, fusion modules, and a   decoder. The Siamese encoder takes a pair of satellite images as input and outputs multi-scale feature maps $\hat X^i_{pre}$ and $\hat X^i_{post}$ at its four stages $i=1,2,3,4$. At each stage, the feature maps are initially down-sampled through a patch embedding layer and then passed through a series of encoder blocks. The structure of our encoder block is shown in Fig.~\ref{fig_overall_architecture}-(b) which comprises our novel efficient local-global context aggregator (ELGCA) module and a convolutional MLP layer.

The focus of our design is the introduction of an efficient local-global context aggregator (ELGCA) module within the encoder block. Our ELGCA module aims to leverage the benefits of both convolution and self-attention \cite{vaswani2017attention} to aggregate local and global context, thereby improving the accuracy of predicted change maps while reducing the number of parameters and FLOPs. 
%We propose a novel ELGCA module within an encoder block to effectively capture the semantic changes between the input image pair ($I_{pre}$ and $I_{post}$).
At each stage,  similar to the base framework, the feature pairs from the encoder blocks are fused through   fusion modules performing simple linear projection,  feature concatenation, and a $1\times 1$ convolution operation to reduce the number of channels. % (see Fig.~\ref{fig_overall_architecture}-(c)). 
These fused features are subsequently passed to a   decoder for change map prediction.

\begin{figure}[!t]
\begin{center}
{\includegraphics[width=1\linewidth, keepaspectratio] {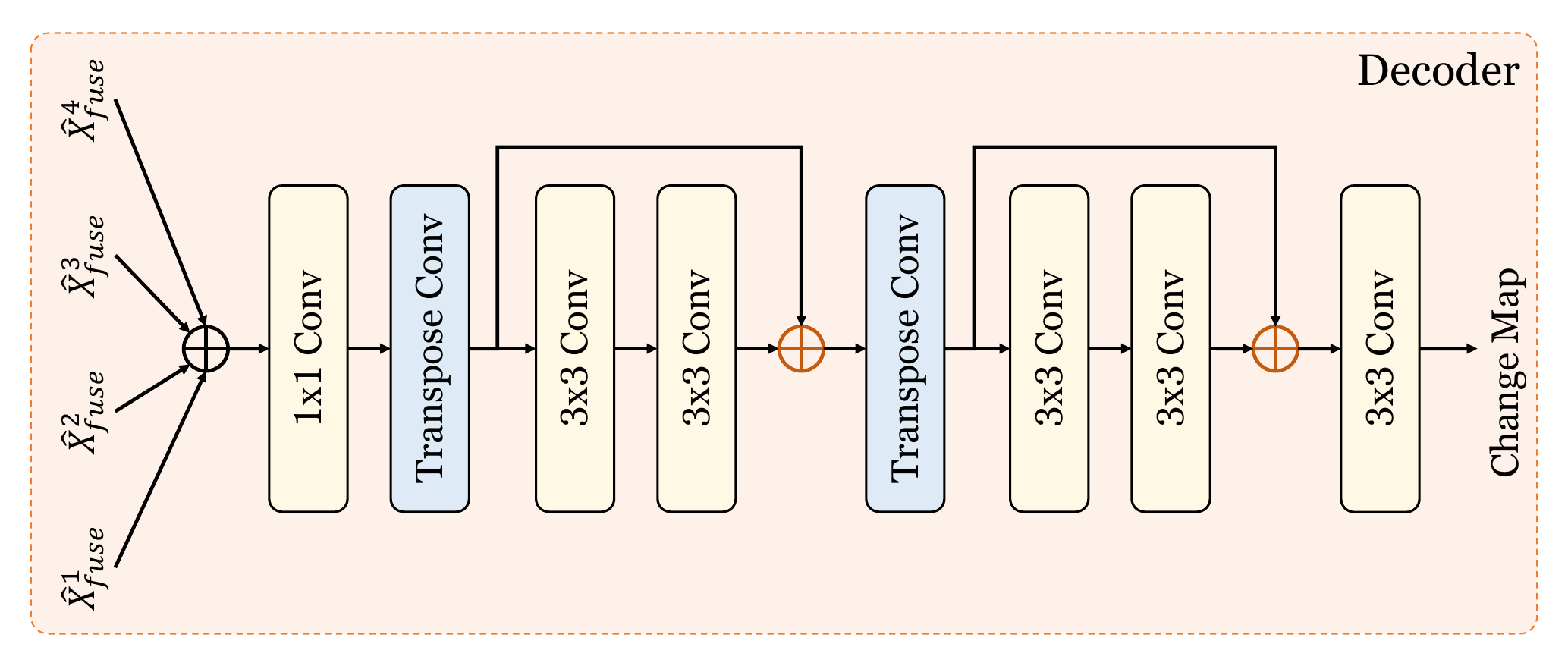}}
\end{center}
\caption{ {ELGC-Net decoder takes fused features $\hat X^{i}_{fused}$ from the four stages (where $i = {1,2,3,4}$) and concatenates them along the channel dimension. Then, it utilizes $1 \times 1$ convolution to project features. Afterward, the upsampling is performed twice to obtain the same spatial dimension as the model input using cascaded transpose convolution and a residual block composed of two convolution layers. %Further upsampling is performed by using the transpose convolution and residual block to obtain the same spatial dimension as the model input. 
Finally, a convolution layer is utilized to obtain the prediction scores having two channels.}}
\vspace{-1.2em}
\label{fig:decoder_architecture}
\end{figure}

\noindent\textbf{Decoder Architecture:}  {Similar to the baseline, our ELGC-Net utilizes a decoder composed of several convolution and transpose convolution layers as illustrated in Fig.~\ref{fig:decoder_architecture}. The multi-scale fused features ($\hat X^{i}_{fused}$ where $i \in {1,2,3,4}$) from four stages are concatenated along channel dimension and fed to a $1 \times 1$ convolution layer. Afterwards, we apply the transpose convolution to increase the spatial resolution of the feature maps. Then, a residual block comprising of two $3 \times 3$ convolution layers is utilized to enhance the feature maps. The transpose convolution, cascaded with a residual block, is utilized twice to obtain the same spatial dimension as the input of the model. Finally, we apply a convolution layer to obtain prediction scores consisting of two channels where the first channel corresponds to no-change and the second channel corresponds to change class scores. The binary change map is obtained by using the \textit{argmax} operation along the channel dimension.}

 Next, we present a detailed discussion of our ELGCA module. 

%############################################
%############################################

\subsection{Efficient Local-Global Context Aggregation (ELGCA) Module} \label{sec_elgca} 
%In this work, we strive to simultaneously achieve both these objectives by replacing the standard self-attention in the encoder with the proposed  ELGCA layer.
Our ELGCA module strives for capturing local and global context information while reducing the computational complexity compared to self-attention through  careful design choices.  
Motivated by the inception module \cite{szegedy2015going} that  stacks multiple convolution operators of different  kernel sizes  in parallel, we split the channels into different groups and perform diverse operations on each channel split (see Fig. \ref{fig_overall_architecture}-(c)).  
Specifically, we channel-wise split the input features and input them into two separate context aggregators to  obtain local  and global contextual information. 
%To do so, we propose to decompose the input features $F^i$ into three contextual levels.
%We obtain the feature representations at three contextual levels.
Let  $X^i \in \mathcal{R}^{H^i \times W^i \times C^i}$ be the input feature to the ELGCA module at  the $i^{th}$ stage, where ($H^i, W^i, C^i$) denote its height, width, and channels, respectively.  We first perform channel-wise splitting of  input feature $X^i$, resulting in  $ X^i_{gl}$,   $X^i_{lo}$ $\in \mathcal{R}^{H^i \times W^i \times \frac{C^i}{2}}$ from which  inputs to our  PT attention  and local context aggregator are obtained. \textit{i.e,}

% by half and obtain the inputs for the local and global contextual aggregators as follows:

\begin{equation}
    X^i_{gl}, X^i_{lo} = Split(X^i)
\end{equation}
% where,  $ X^i_{gl}$, $X^i_{lo}$ $\in \mathcal{R}^{H^i \times W^i \times \frac{C^i}{2}}$.\\
%\noindent  % The advantages of employing depthwise convolution are in three folds: (1) the model can capture effective local contextual information and (2) it minimizes the number of model parameters and FLOPs. (3) The underlying  attributes of the depthwise convolution  diminish the requirement for conventional position embedding \cite{chu2021conditional}, known as tokenization in ViTs \cite{dosovitskiy2020image}.
Next, we describe the PT attention and local  context aggregation within our ELGCA module.  

\noindent \textbf{Pooled-transpose (PT)  Attention:} 
As mentioned earlier, the  standard self-attention has  quadratic computational complexity with respect to number of tokens, whereas the proposed PT attention has only linear complexity with respect to tokens. % in our ECGCA module.    
%To extract the features at the global contextual level, 
We perform   a $1 \times 1$ convolution 
 on  $X^i_{gl}$  and split  it into $Z^i$, $Q^i$, $K^i$, and $V^i$  features, respectively. Here, $Q^i$, $K^i$, and $V^i$ are used as \textit{query}, \textit{key}, and \textit{value} for our PT attention capturing global contextual information, whereas  $Z^i$ performs   multi-channel feature aggregation. The key steps of our PT attention are  the following: 

%Therefore, we propose pooled-transpose (PT) attention to mitigate the issues associated with self-attention operations. 
At first, our PT attention   performs average- and max- pooling operations over the  $Q^i$ and $K^i$ features, respectively to obtain pooled features $\bar{Q}^i$ and 
$\bar{K}^i$ that are robust to slight  variations.   Specifically, we perform  $3\times 3$  average (Avg) pooling with stride 2 on  $Q^i$ to capture  mean features $\bar {Q}^i$,  and employ a  $2\times 2$  max (Max) pooling with stride 2 on  $K^i$ to obtain  dominant features $\bar {K}^i$. 

% features are passed through the average (Avg) pooling layer (with kernel size 3 and stride 2) whereas $K^i$ features are input to the maximum (Max) pooling layer (with kernel size 2 and stride 2).  The aim is to obtain the dominant  features for the  $K^i$ matrix

Secondly, we employ transposed attention ($G$), between    $\bar {Q}^i$,  $\bar {K}^i$, and $V^i$ embedding,  having linear complexity with respect to the number of tokens on a subset of feature channels ($C/4$). Our PT attention operations  to obtain  feature representation $A^i_{att}$   can be summarized as follows:

\begin{equation}
    {A^i}_{att} = V^i  \times [\sigma (\bar{K^i}^T  \times \bar{Q}^i)],    
\end{equation}

where $\times$ indicates  matrix multiplication   and $\sigma$ represents the Softmax operations, respectively. Here, $\bar {Q}^i$, $\bar{K}^i$  $\in \mathcal{R}^{\frac{H^i}{2}\frac{W^i}{2} \times \frac{C^i}{4}}$, and  ${V}^i$ $\in \mathcal{R}^{H^iW^i \times \frac{C^i}{4}}$. 
Our PT attention leads to improved performance and generalization capabilities,  while simultaneously reducing the convergence time and computational cost. 

\noindent\textbf{Local Contextual Aggregation: }
In order to capture spatially and channel-wise local contextual information, we perform  $3\times3$  depthwise convolution on input $X^i_{lo}$, resulting in feature representations $\bar X^i_{lo}$. This   depthwise convolution based local context aggregation effectively captures local contextual information with the minimum  number of model parameters and FLOPs, and it alleviates the requirement of positional embedding \cite{chu2021conditional}. 

Finally, the local and global context aggregated features  with various  receptive fields (such as $\bar X^i_{lo}$, $Z^i$, and $A^i_{att}$) are merged using concatenation operation to  obtain enriched local-global contextual aggregated feature ($\bar X^i$). In contrast to \cite{xiao2021early, tu2022maxvit, fiaz2023sat}, where context aggregators are employed in a sequential manner,  our ELGCA module exploits multiple context aggregators at different subsets of the channels in a parallel fashion which leads to reduced computational complexity.   
%\begin{flalign}
%\begin{aligned}
%    \bar X_{inter} = GeLU(Conv(LN(\bar X))), \\
%    \tilde X = Conv(  GeLU(DWConv(\bar X_{inter}))  \oplus \bar X_{inter}),    
%\end{aligned}
%\end{flalign}

In summary, our ELGC-Net CD framework employs the ELGCA module at different stages of the encoder block and generates pre- and post-change feature maps ($\hat X^i_{pre}$ and $\hat X^i_{post}$) from the two streams. These $\hat X^i_{pre}$ and $\hat X^i_{post}$ features at each stage are fused ($\hat X^i_{fuse}$) using a fusion module and then taken as input to a decoder for obtaining the final change map predictions.

%#############################
\begin{table*}[t!]
\centering
\caption{State-of-the-art comparison on LEVIR-CD, DSIFN-CD, and CDD-CD. The results are reported in terms of the change class IoU, change class F1, and overall accuracy (OA) metrics. Our ELGC-Net efficiently combines the global and local contextual features and achieves superior performance in terms of all metrics over these datasets. Here, the best two results are highlighted in \textcolor{red}{red} and \textcolor{blue}{blue} colors, respectively.}
\label{tbl:comaprison_on_LEVIR_DSIFN_CDD}
\setlength{\tabcolsep}{12pt}
\scalebox{0.94}{
\begin{tabular}{|l|c|c|c|c|c|c|c|c|c|c|c|c|c|} \hline
\multicolumn{1}{|l|}{\multirow{2}{*}{Method}} & \multicolumn{1}{c|}{\multirow{2}{*}{Params (M)}} &  \multicolumn{3}{c|}{LEVIR-CD} & \multicolumn{3}{c|}{DSIFN-CD} & \multicolumn{3}{c|}{CDD-CD}\\ 
\cline{3-11} 
\multicolumn{1}{|l|}{}
& \multicolumn{1}{c|}{}
& IoU & F1  & OA  
& IoU & F1  & OA 
& IoU & F1  & OA  \\ 
%\cline{1-1}
\hline 
\hline

BIT  \cite{chen2021_bit}
& 18.78 
& 80.68 & 89.31  & 98.92  
%& 52.97 & 69.26 & 89.41  \\ 
& 51.22 & 67.74 & 89.72  
& 80.01 & 88.90 & 97.47  \\

FC-Siam-diff \cite{daudt2018_fcsiam}
& -- %1.35 
%& -- & --
& 75.92 & 86.31 & 98.67  
%& 45.50 & 62.54 & 86.63  \\
& 48.44 & 65.26 & 89.06  
& 54.57 & 70.61 & 94.95  \\

FC-Siam-conc \cite{daudt2018_fcsiam}
& -- %1.55
%& -- & --
& 71.96 & 83.69 & 98.49 
%& 42.56 & 59.71 & 87.57 \\
& 44.33 & 61.43 & 84.51 
& 60.14 & 75.11 & 94.95 \\

DTCDSCN \cite{liu2020building} 
& 41.07
& 78.05 & 87.67 & 98.77
%& 49.76 & 63.72 & 84.91 \\
& 48.46 & 65.29 & 88.14 
& 85.34 & 92.09 & 98.16 \\

%DASNet \cite{chen2020dasnet} 
%& 108.69 
%& -- & --
%& 74.65 & 79.91 & 94.32
%& -- & -- & -- 
%& -- & 92.70 & 98.00 \\

SNUNet \cite{fang_snunet} 
& 12.03 
& 78.83 & 88.16 & 98.82
%& 71.67 & 83.50 & -- \\ %98.71 \\
& 47.13 & 64.07 & 86.54  %98.71 \\
& 72.11 & 83.89 & 96.23 \\

STANet \cite{chen2020spatial} 
& 16.93 
& 77.40 & 87.30 & 98.66
%& 47.66 & 64.56 & 88.49 \\
& 45.12 & 62.19 & 85.00 
& 72.22 & 84.12 & 96.13 \\

H-TransCD \cite{ke2022hybrid} 
& 183.83 
& 81.92 & 90.06 & 99.00
& -- & -- & -- 
& -- & -- & -- \\

IFNet   \cite{zhang2020deeply} 
& 35.99 
& 78.77 & 88.13  & 98.87 
& 43.28 & 60.42 & 84.41 
& 71.91 & 90.30 & 97.71 \\

MSTDSNet   \cite{song2022mstdsnet} 
& -- 
& 78.73 & 88.10  & 98.56 
& -- & -- & -- 
& -- & -- & -- \\

TransUNetCD \cite{li2022transunetcd}
& 95.49 
& \textcolor{blue}{83.67} & \textcolor{blue}{91.11} & -- 
%& 57.95 & 66.62 & -- \\
& -- & -- & -- 
& \textcolor{red}{94.50} & \textcolor{red}{97.17} & -- \\

ChangeFormer  \cite{changeformer} 
& 41.03 
& 82.48 & 90.40  & 99.04 
%& 76.48 & 86.67 & 95.56  \\
& 53.26 & 69.50 & 90.56  
& 81.53 & 89.83 & 97.68 \\

\hline

%Baseline 
%& 81.93 & 90.07 & 99.00 
%&  \textcolor{blue}{85.45} & \textcolor{blue}{92.15} & \textcolor{blue}{97.35}  \\

\hline
\textbf{ELGC-Net-LW (Ours)} 
& \textcolor{red}{6.78} 
& 82.36 & 90.33 & 99.03 
%& \textcolor{blue}{84.29} & \textcolor{blue}{91.48} & \textcolor{blue}{97.06} \\ 
& \textcolor{blue}{57.92} & \textcolor{blue}{73.35} & \textcolor{blue}{91.57} &
\textcolor{blue}{93.48} & \textcolor{blue}{96.63} & \textcolor{blue}{99.21} \\ 

%\hline
\textbf{ELGC-Net (Ours)} 
& \textcolor{blue}{10.57} 
& \textcolor{red}{83.83} & \textcolor{red}{91.20}  &  \textcolor{red}{99.12} 
%& \textcolor{red}{86.03} &  \textcolor{red}{92.49} &  \textcolor{red}{97.45} \\
& \textcolor{red}{58.42} &  \textcolor{red}{73.76} &  \textcolor{red}{92.68} &
\textcolor{red}{94.50} &  \textcolor{red}{97.17} &  \textcolor{red}{99.33} \\ 

\hline

\end{tabular}}
\end{table*}

%################################################
\begin{figure*}[!t]
\begin{center}
{\includegraphics[width=1\linewidth, keepaspectratio] {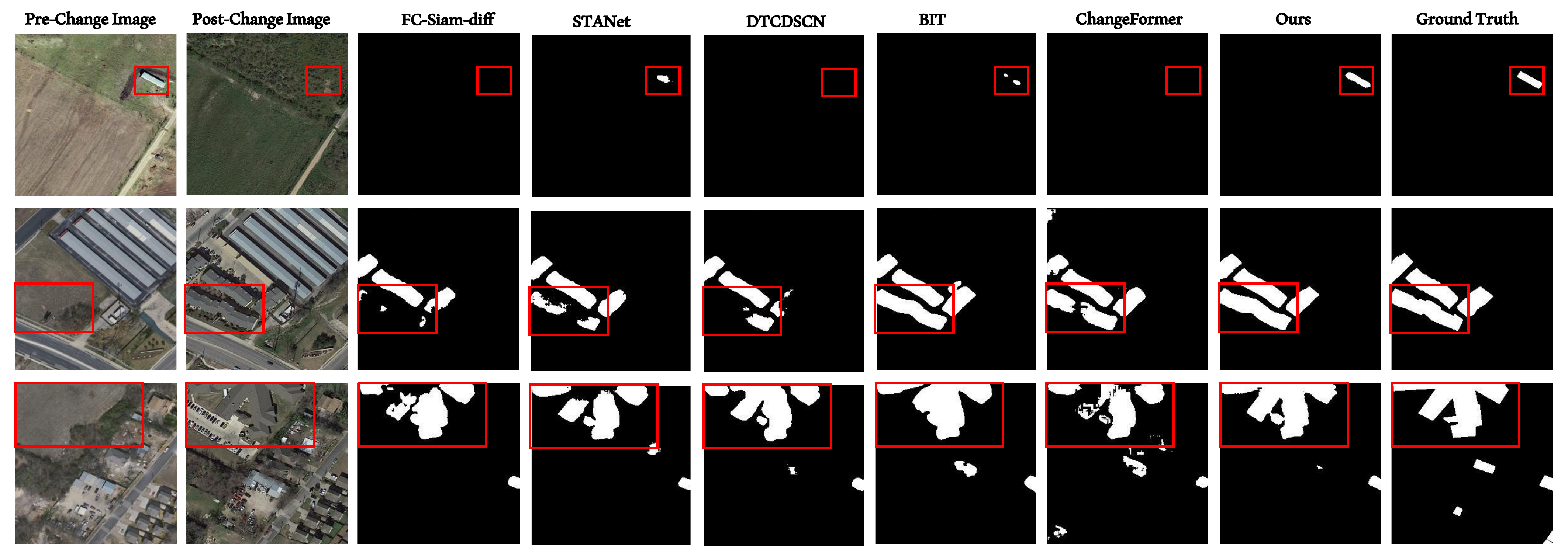}}
\end{center}
\caption{ {Qualitative results on the LEVIR-CD dataset. We present comparison with the best five existing change detection methods in literature, whose codebases are publicly available. The highlighted region shows that our method (ELGC-Net) is better at detecting the change regions as compared to  CNN-based including FC-Siam-diff \cite{daudt2018_fcsiam}, STANet \cite{chen2020spatial}, DTCDSCN \cite{liu2020building} and transformer-based such as BIT \cite{chen2021_bit} and ChangeFormer \cite{changeformer} methods.}}
\label{fig_levir_vis}
\end{figure*}
%################################################

\section{Experiments}

\subsection{Datasets and Evaluation Protocols}
In our experiments, we use three public remote sensing change detection datasets to verify the effectiveness of the proposed network. 

\subsubsection{LEVIR-CD \cite{chen2020spatial}} is a large-scale building change detection dataset. It consists of 637 high-resolution (0.5 m/pixel) image pairs of spatial size $1024 \times 1024$ taken from Google Earth. The time span of these image pairs ranges from 5 to 14 years. The dataset focuses on the changes related to building construction and demolition. In our experiments, we used the cropped version of the dataset having non-overlapping cropped patches of size $256 \times 256$. Following ~\cite{changeformer}, we utilize the default data split of size 7120, 1024, and 2048 image pairs for train, validation and test sets respectively.

\subsubsection{DSIFN-CD \cite{zhang2020deeply}} is another binary change detection dataset that is collected from Google Earth. The dataset contains 6 high-resolution bi-temporal images covering six cities of China. The dataset is available in cropped sub-image pairs of size $256 \times 256$ with train, validation, and test split of 14400, 1360, and 28 image pairs, respectively.

\subsubsection{CDD-CD \cite{lebedev2018_cdd}} is a well-known remote sensing change detection dataset that contains 11 seasonal varying image pairs. Among these, 7 image pairs have a spatial size of $4725 \times 2700$ pixels while 4 image pairs are of size $1900 \times 1000$ pixels. The dataset has been used in the clipped form (\cite{chen2020dasnet, li2022transunetcd}) of size $256 \times 256$ pixels and contains 10000, 3000, and 3000 image pairs in train, validation, and test sets respectively.

\noindent\textbf{\textit{Evaluation Protocol:}}
Similar to the ~\cite{changeformer}, we evaluate the change detection results  using  \textit{change class} intersection over union (IoU), \textit{change class} F1-score, and overall accuracy (OA) protocols on all the datasets.
%The \textit{change class} IoU is the challenging metric as a major portion of the regions in the image pairs belong to the background class.

%################################################
\begin{figure*}[t!]
\begin{center}
{\includegraphics[width=1\linewidth, keepaspectratio] {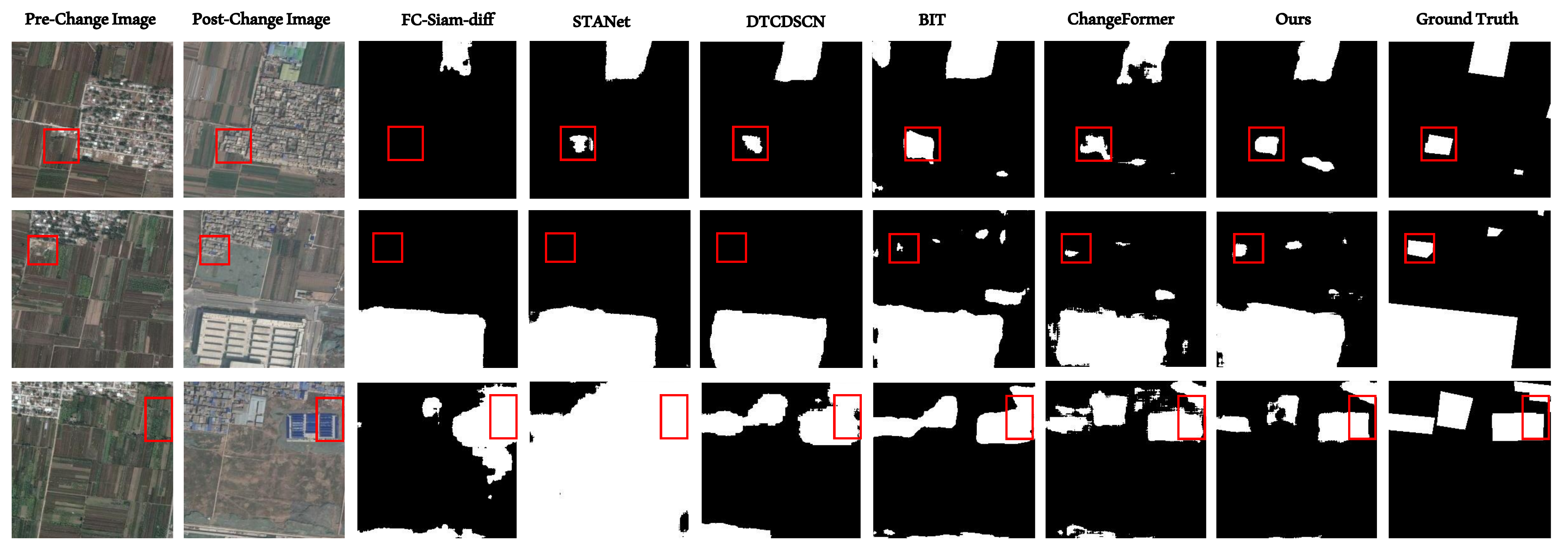}}
\end{center}
\caption{ {Qualitative results on the DSIFN-CD dataset with a comparison presented with the best five existing change detection methods in the literature, whose codebases are publicly available. The highlighted region in red box shows that ELGC-Net is better to identify the change regions as compared to the CNN-based FC-Siam-diff \cite{daudt2018_fcsiam}, STANet \cite{chen2020spatial}, DTCDSCN \cite{liu2020building} as well as transformer-based BIT \cite{chen2021_bit} and ChangeFormer \cite{changeformer} methods.}}
\label{fig_dsifn_vis}
\end{figure*}
%################################################

\subsection{Implementation Details}
The encoder of the proposed ELGC-Net is composed of four stages having 3,3,4,3 encoder blocks and outputs feature maps that have channel dimensions of 64, 96, 128, and 256, respectively. The input to the encoder is a pair of co-registered images with the size of $256 \times 256 \times 3$. The decoder outputs the binary change map of the same spatial resolution.
%During training, pixel-wise cross-entropy loss is used to measure the performance of the network. 
Furthermore, random flip, scale crop, color jitter, and Gaussian blur data augmentations are used in the training process. The network is trained on 4 Nvidia A100 GPUs and model weights are randomly initialized. We set the learning rate equal to 3.1e-4 and train the model for 300 epochs. The value of the learning rate is selected empirically. We linearly decay the learning rate till the last epoch. Following \cite{changeformer}, we use AdamW optimizer having weight decay of 0.01 and beta values of (0.9, 0.999).

During inference, we freeze the model weights and pass the input image pair to the model. The network outputs the probability change map having two channels. A binary change map is obtained through the argmax operation along the channel dimension.

\noindent\textbf{Loss Function: } During training, we utilize pixel-wise cross-entropy loss to measure the performance of the network. For binary change detection problem, cross-entropy loss function is mathematically expressed as:
\begin{equation}
    Loss = -\frac{1}{N}\sum_{i=1}^{N} {y_i}  {log(\hat {y}_i)} + {(1 - y_i)}  {log(1 - \hat {y}_i)}   
\end{equation}

where $y$ is the ground truth class label, $\hat y$ is the prediction probability and $N$ is the number of pixels.
%######################################################
\begin{figure*}[t!]
\begin{center}
{\includegraphics[width=1\linewidth, keepaspectratio] {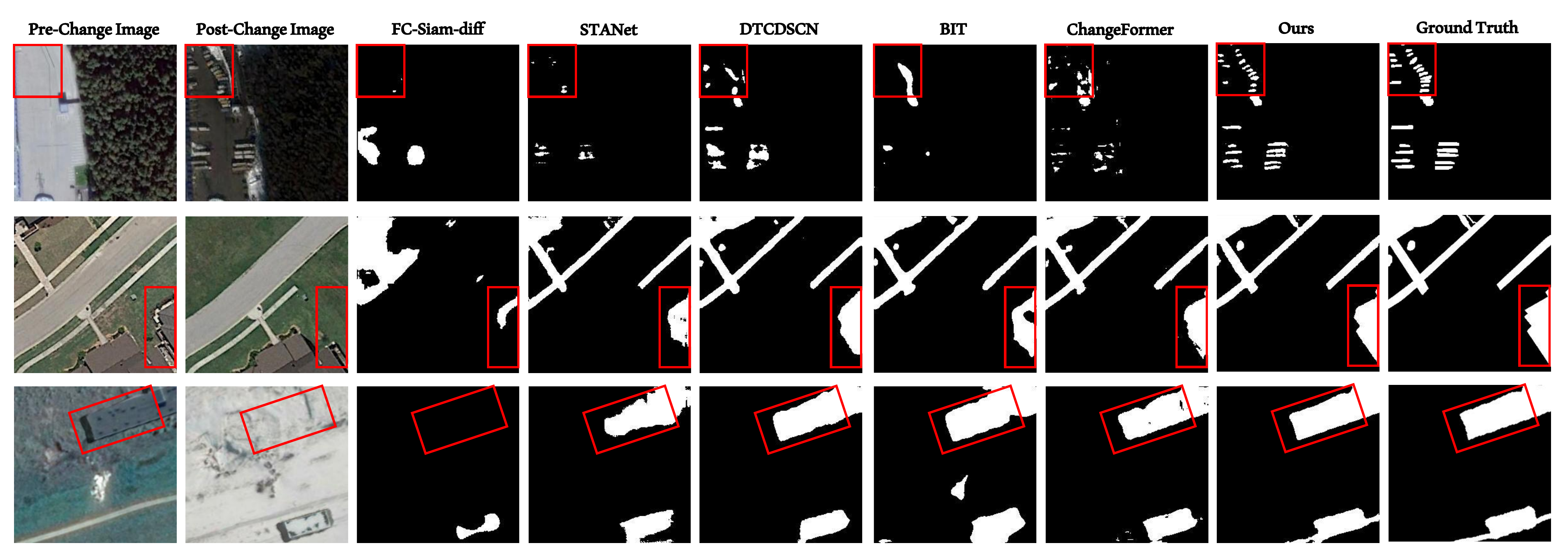}}
\end{center}
\caption{ {Qualitative results on the CDD-CD dataset. We show the comparison with the best five existing change detection approaches in the literature, whose codebases are publicly available. The highlighted region in the red box shows that ELGC-Net is better to detect the change regions with clear boundaries between the pre- and post-change images.}}
\label{fig_cdd_vis}
\end{figure*}
%######################################################

%##############################
\begin{table*}[t!]
\centering
\caption{
 {
Ablation study over the  LEVIR-CD. We present the impact of our contributions including local contextual aggregation, multi-channel aggregation, and PT attention (global contextual aggregation) by replacing the self-attention of the baseline. The local contextual aggregation using depthwise convolution (row 1) has the least parameters. 
PT attention (row 3) presents a significant improvement over local contextual aggregation  (row 1) and multi-channel aggregation (row 2) in terms of IoU while having increased parameters. 
Our final approach (row 7) aggregates all contextual information results in optimal IoU score while having comparable parameters 
to local contextual aggregation  (row 1). The best results are shown in bold text.}}
\label{tbl:comp_of_features}
\setlength{\tabcolsep}{9.5pt}
\scalebox{1.1}{
\begin{tabular}{|c|c|c|c|c|c|c|} \hline
\multicolumn{3}{|c|}{Context Aggregators} &  \multirow{2}{*}{Params (M)} & \multirow{2}{*}{IoU} & \multirow{2}{*}{F1} & \multirow{2}{*}{OA} \\
% & DSIFN-CD \\
\cline{1-3}
%\cline{6-6}

Local Contextual Aggregation%$(3 \times 3)$ DW-Conv 
& Multi-channel Aggregation %$(1 \times 1)$ Conv 
& PT Attention  &  & & & \\ 
\hline
\hline
%\xmark &  \xmark & Baseline & \textbf{11.63} & \textbf{138557.9} & 81.93 \\ % & 85.45 \\
$\checkmark$ &  \xmark & \xmark & \textbf{10.43} &  81.69 & 89.93 & 98.98\\ % & 83.78 \\
%$\checkmark$ &  \xmark & \xmark & \textbf{10.43} & \textbf{123476.9} & 82.69 \\ % & 84.78 \\
\xmark & $\checkmark$ & \xmark & 10.72 &  81.24  & 89.65 & 98.97 \\ % & 83.26 \\
%\xmark & $\checkmark$ & \xmark & 10.72 & 123670.2 & 83.24\\ %  & 85.26 \\
\xmark & \xmark & $\checkmark$ & 11.32  & 82.55  & 90.44 & 99.02 \\ % & 83.16 \\
%$\checkmark$ & $\checkmark$ & \xmark & 10.50 & 123519.6 & 83.18 \\ % & 84.35 \\
$\checkmark$ & $\checkmark$ & \xmark & 10.50 & 82.18  & 90.21 & 99.00 \\ % & 84.35 \\
$\checkmark$ & \xmark & $\checkmark$ & 10.65 & 83.13   & 90.79 & 99.08 \\ % & 85.94 \\
\xmark & $\checkmark$ & $\checkmark$ & 11.02 & 82.72  & 90.54 & 99.04 \\ %  & 84.86 \\
$\checkmark$ & $\checkmark$ & $\checkmark$ & 10.57 & \textbf{83.83}  & \textbf{91.20} & \textbf{99.12} \\ %  & \textbf{86.03} \\
\hline
\end{tabular}
} 
\end{table*}
%##############################

%##############################
\begin{table*}[]
\centering
\caption{ { Comparison of parameters, FLOPs, memory consumption, and change class IoU metric on LEVIR-CD. ELGC-Net has fewer parameters and FLOPs as compared to the baseline and ChangeFormer.  The best results are shown in bold text.}}
\label{tbl:comp_of_params}
\setlength{\tabcolsep}{15pt}
%\setlength{\extrarowheight}{1pt}
%\adjustbox{width=\columnwidth}{
\scalebox{1.1}{\begin{tabular}{|l|c|c|c|c|c|c|} %\toprule[0.08em]
\hline
Method & Params (M) & FLOPs (M) & Memory (MBs) & IoU  & F1 & OA\\ 
%\toprule[0.08em]
\hline
\hline
ChangeFormer \cite{changeformer} & 41.03 & 138803.4 & 989 & 82.48 & 90.40 & 99.04 \\
Baseline & 11.63 & 138557.9 & 2087 & 81.93 & 90.07 & 99.00 \\
{ELGC-Net (SA)\textsuperscript{\textdagger} } & {10.57} & 127133.5 & 1540 & 83.10 & 90.77 & 99.08 \\ \hline
ELGC-Net-LW (Ours) & \textbf{6.78} & \textbf{19815} &  \textbf{704} & {82.36} & 90.33 & 99.03\\
%774
ELGC-Net (Ours) & {10.57} & {123590.5} & 722.5 & \textbf{83.83} & \textbf{91.20} & \textbf{99.12} \\
\hline
\multicolumn{4}{l}{
\textdagger \footnotesize{ represents PT attention is replaced with self-attention.}}
\end{tabular}
}
\end{table*}

%##############################

\subsection{State-of-the-art Quantitative Comparison}
We present a quantitative comparison of our approach with state-of-the-art (SOTA) methods over LEVIR-CD, DSIFN-CD, and CDD-CD datasets in Table \ref{tbl:comaprison_on_LEVIR_DSIFN_CDD}.% and   over CDD-CD dataset in Table \ref{tbl:comaprison_on_CDD}, respectively.

Tab.~\ref{tbl:comaprison_on_LEVIR_DSIFN_CDD} shows that BIT \cite{chen2021_bit}, H-TransCD \cite{ke2022hybrid}, ChangeFormer \cite{changeformer}, and TransUNetCD \cite{li2022transunetcd} perform better on LEVIR-CD dataset by achieving IoU score of 80.68\%, 81.92\%, 82.48\% and 83.67\% respectively. 
%Similarly, our baseline method achieves the same results as H-TransCD and comparable results to ChangeFormer attaining an IoU score of 81.93\%. 
Our ELGC-Net achieves superior performance in terms of all metrics against the compared methods while obtaining the IoU gain of 1.4\% and 0.2\% over ChangeFormer \cite{changeformer} and TransUNetCD \cite{li2022transunetcd}, respectively. Noted, our method surpasses the state-of-the-art methods without using a pre-trained backbone.
Furthermore, we introduce ELGC-Net-LW which comprises a lightweight decoder. The aim is to further reduce the trainable parameters as the decoder is performing operations on high-resolution feature maps.
%ELGC-Net-LW utilizes a combination of depth-wise and standard convolution operations as well as replace the transpose convolution with a bilinear interpolation operation.
ELGC-Net-LW replaces the transpose convolution with bilinear interpolation operation and utilizes a combination of depthwise and standard convolution operations in place of computationally expensive residual modules.
Despite having fewer trainable parameters, our lightweight ELGC-Net-LW also performs better compared to SOTA methods except for TransUNetCD and ChangeFormer which have slightly better IoU scores. However, TransUNetCD and ChangeFormer have significantly higher numbers of trainable parameters as listed in Tab.~\ref{tbl:comaprison_on_LEVIR_DSIFN_CDD}.

Similar to the LEVIR-CD, our method demonstrates favorable performance over the DSIFN-CD dataset as shown in Tab. \ref{tbl:comaprison_on_LEVIR_DSIFN_CDD}.
%The ChangeFormer, transformer-based best performing,  gives the IoU score of 53.26\%. Our ELGC-Net achieves absolute gains of 9.55\%, 5.82\%, and 1.89\% in terms of IoU, F1, and overall accuracy (OA)
 We notice that our ELGC-Net achieves favorable performance against the state-of-the-art approaches.
{It is worth mentioning that ELGC-Net-LW, having very few parameters and without using any pre-trained backbone, performs outstandingly on DSIFN-CD.} %and achieves the IoU score of 84.29\% on DSIFN-CD.}
%Even the selected baseline achieves significantly better performance than the best existing methods as shown in Tab.~\ref{tbl:comaprison_on_LEVIR_DSIFN}. 

We also report the state-of-the-art comparison over the CDD-CD dataset in  Tab.~\ref{tbl:comaprison_on_LEVIR_DSIFN_CDD}.
We observe that CNN-based DTCDSCN \cite{liu2020building} obtains a 85.34\% IoU score. However, our method achieves a significant gain and provides a 94.50\% score of IoU.
On the other hand, transformer-based TransUNetCD \cite{li2022transunetcd} provides the best performance on CDD-CD by achieving an IoU score of 94.50\%. Our ELGC-Net is also obtaining the same score compared to the TransUNetCD \cite{li2022transunetcd}. Notably, our ELGC-Net does not use any pre-trained backbone feature extractor and still achieves the same results as the state-of-the-art method. This demonstrates the proficiency of our proposed efficient context aggregator to better capture the local and global contextual information for the CD task.

\subsection{Qualitative Comparison}
We present the qualitative results of ELGC-Net in Fig.~\ref{fig_levir_vis}, Fig.~\ref{fig_dsifn_vis}, and Fig.~\ref{fig_cdd_vis} for LEVIR-CD, DSIFN-CD, and CDD-CD datasets, respectively. Fig.~\ref{fig_levir_vis} shows that 
 {
ELGC-Net is able to detect the small change (see first row) that is not detected by CNN-based FC-Siam-diff \cite{daudt2018_fcsiam}, STANet \cite{chen2020spatial}, DTCDSCN \cite{liu2020building} as well as transformer-based BIT \cite{chen2021_bit} and ChangeFormer \cite{changeformer} methods. Similarly, in the second and third rows, our method accurately detects the change
regions while ChangeFormer \cite{changeformer} and BIT \cite{chen2021_bit} methods strive to identify the changes precisely.}

In Fig.~\ref{fig_dsifn_vis}, our ELGC-Net is able to detect the true change regions while suppressing the false detections whereas  { both CNN-based and transformer-based including  FC-Siam-diff \cite{daudt2018_fcsiam}, STANet \cite{chen2020spatial}, DTCDSCN \cite{liu2020building}, BIT \cite{chen2021_bit} and ChangeFormer \cite{changeformer} have more false change detections in addition to the true change regions.} The reason is that our ELGC-Net has better capability to capture the discriminative features at local and global context levels and possesses the properties of both convolution and self-attention operations that complement each other for the detection of subtle and large change regions.

 {Fig.~\ref{fig_cdd_vis} shows the performance comparison among  CNN-based  including FC-Siam-diff \cite{daudt2018_fcsiam}, STANet \cite{chen2020spatial}, DTCDSCN \cite{liu2020building} and  transformer-based  such as BIT \cite{chen2021_bit} and ChangeFormer \cite{changeformer}  with our ELGC-Net over the CDD-CD dataset. We see that the proposed ELGC-Net is capable of detecting subtle changes considerably better than the SOTA method (first row). Furthermore, the BIT \cite{chen2021_bit} and ChangeFormer \cite{changeformer} approaches struggle to detect the change region accurately in the bottom-right of the second row which is precisely detected by our ELGC-Net.}

\subsection{Ablation Study}

\subsubsection{Ablation of contextual features in ELGCA module}

We evaluate  the effectiveness of our contributions including  local contextual aggregation, multi-channel aggregation, and PT attention (global contextual aggregation) over the LEVIR-CD dataset. To do so, we iteratively apply different combinations of our contributions and present their parameters, 
%FLOPs, 
and IoU scores  as shown in Tab.~\ref{tbl:comp_of_features}.
%First, we apply individual contributions within the baseline and present their parameters and FLOPs.
Although the local contextual information using depthwise convolutions has minimum parameters (10.43M) 
%and FLOPs (123476.9M)
compared to other variants ( as illustrated in Tab.~\ref{tbl:comp_of_features} row 1), it has reduced  IoU score of 81.69\%. Whereas PT attention has an IoU score of 82.55\% while having increased parameters. 
%and FLOPs.
Nevertheless, our final network aggregating all the contextual information provides the best IoU score of 83.83\%, with comparable  parameters 
%and FLOPs 
compared to row 1 as displayed in the last row in Tab.~\ref{tbl:comp_of_features}. It is evident that our final ELGCA module exhibits reduced parameters and better IoU score while leveraging the aggregation of different context aggregators in a parallel fashion.

%Each branch is contributing to the performance of the network as depicted in Tab.~\ref{tbl:comp_of_features}. The network performs slightly worse when we remove the $1 \times 1$ convolution features and self attention layer resulting in decreased IoU score, however, removing self attention layer minimizes the number of trainable parameters and FLOPs. The performance of the network degrades more when we remove the $3 \times 3$ depth-wise convolution branch that is responsible for capturing the local structural details in the image pairs. Finally, combining all the features provide the best IoU score of 83.83\% and 86.03\% for LEVIR-CD and DSIFN-CD datasets, respectively, with a slight increase in the parameters and FLOPs as compared to the first row in Tab.~\ref{tbl:comp_of_features}.

\subsubsection{Comparison of Parameters and FLOPs}
We present the comparison of the parameters and FLOPs of our ELGC-Net with ChangeFormer \cite{changeformer} and baseline. Tab.~\ref{tbl:comp_of_params} shows that ELGC-Net has fewer parameters and FLOPs compared to state-of-the-art ChangeFormer \cite{changeformer} while it outperforms on all three datasets. We also demonstrate the effectiveness of our pooled-transpose (PT) attention 
to have linear complexity with respect to tokens (
as shown in  Tab.~\ref{tbl:comp_of_params} row 3).
To do so, we replace the PT attention with standard self-attention  and observed that our PT attention exhibits fewer FLOPs while enhancing the IoU score. Furthermore,  the peak memory usage during the inference stage also shows that ELGC-Net consumes lesser memory.
{Additionally, we present the results of ELGC-Net-LW that utilizes a light-weight decoder in Tab.~\ref{tbl:comp_of_params}. Although, the IoU score of ELGC-Net-LW is slightly reduced, however, the number of trainable parameters and FLOPs are considerably lower than the baseline and ChangeFormer which further demonstrates the effectiveness of the ELGCA module.
}

%##############################
\begin{table}[]
\centering
\caption{  {Comparison of different attention layers and ELGCA module on LEVIR-CD. The best results are shown in bold text.}}
\label{tbl:comp_of_attn}
%\setlength{\tabcolsep}{8pt}
%\setlength{\extrarowheight}{1pt}
%\adjustbox{width=\columnwidth}{
\scalebox{0.8}{
\begin{tabular}{|l|c|c|c|c|c|} %\toprule[0.08em]
\hline
Layers & Params (M) & FLOPs (M) & IoU & F1 & OA \\ 
%\toprule[0.08em]
\hline
\hline

Self Attention (Baseline) & 11.63 & 138557.9 & 81.93 & 90.07 & 99.00\\
Separable Attention \cite{mehta2022mobilevit} & 11.32 & 124104.8 & 83.13 & 90.78 & 99.07\\
EfficientFormer Attention \cite{li2022efficientformer} & 15.07 & 267518.1 & 81.64 & 89.89 & 98.99\\
ELGCA & \textbf{10.57} & \textbf{123590.5} & \textbf{83.83} & \textbf{91.20} & \textbf{99.12}\\
\hline
\end{tabular}
} 
\end{table}
%##############################

\subsubsection{Comparison between other context aggregators and Our's ELGCA }
In Tab.~\ref{tbl:comp_of_attn}, we present the comparison of the ELGCA module with self-attention, separable attention \cite{mehta2022mobilevit}, and EfficientFormer attention \cite{li2022efficientformer} layer on LEVIR-CD. We replaced the ELGCA module in the proposed network with the self-attention, separable attention, and EfficientFormer attention layer, and train the network with the same hyperparameters. Tab.~\ref{tbl:comp_of_attn} shows that self-attention and separable attention \cite{mehta2022mobilevit} have 11.63 and 11.32 million parameters respectively, and separable attention \cite{mehta2022mobilevit} performs better as compared to self-attention \cite{changeformer} and EfficientFormer attention \cite{li2022efficientformer} by achieving IoU score of 83.13\%. However, the proposed ELGCA module performs remarkably better compared to these attention layers by obtaining an IoU score of 83.83\% while having less number of trainable parameters and FLOPs.

%##############################
\begin{table}[]
\centering
\caption{  {Ablation study of different pooling layers in self-attention of ELGCA module on LEVIR-CD. Average and maximum pooling is applied to the \textit{query} and \textit{key} features of the self-attention before computing the matrix multiplication. An optimal combination of average and maximum pooling gives the best performance highlighted in bold.}}
\label{tbl:ablation_of_pooling}
%\setlength{\tabcolsep}{8pt}
%\setlength{\extrarowheight}{1pt}
%\adjustbox{width=\columnwidth}{
\scalebox{1.2}{
\begin{tabular}{|c|c|c|c|c|} \hline
\multicolumn{2}{|c|}{Pooling applied to} & \multirow{2}{*}{IoU} & \multirow{2}{*}{F1} & \multirow{2}{*}{OA} \\
\cline{1-2}
Query Features & Key Features &  &  & \\ 
\hline
\hline
- & - & 82.80 & 90.59 & 99.05 \\
Avg Pool & Avg Pool & 83.30 & 90.89 & 99.08 \\
Max Pool & Max Pool & 83.00 & 90.71 & 99.06 \\
Max Pool & Avg Pool & 83.24 & 90.85 & 99.08\\
\textbf{Avg Pool} & \textbf{Max Pool} & \textbf{83.83} & \textbf{91.20} & \textbf{99.12} \\
\hline
\end{tabular}
} 
\end{table}
%##############################

%##############################
\begin{table}[]
\centering
\caption{  {Comparison of loss functions on LEVIR-CD applied during the training of our model. The best results are highlighted in bold text.}}
\label{tbl:comp_of_losses}
\setlength{\tabcolsep}{11pt}
\scalebox{1.2}{
\begin{tabular}{|l|c|c|c|} 
\hline
Loss Function & IoU & F1 & OA \\ 
\hline
\hline
Focal Loss  & 79.27 & 88.44 & 98.84\\
mIoU Loss  & 77.51 & 87.32 & 98.75\\
Cross-Entropy Loss  & \textbf{83.83} & \textbf{91.20} & \textbf{99.12} \\
\hline
\end{tabular}
} 
\end{table}
%##############################

\subsubsection{Ablation of pooling layers in ELGCA}
Fig.~\ref{fig_overall_architecture}-(c) shows the architecture of the proposed ELGCA module. As discussed earlier in the ELGCA module,  First, average and maximum pooling operations are applied on the \textit{query} and \textit{key} features to obtain slightly robust features against minor variations. Secondly, we employ the transposed attention having linear complexity with respect to tokens, to reduce the computational cost of matrix multiplication. We conduct various experiments with different pooling layers and present the results in Tab.~\ref{tbl:ablation_of_pooling}. 
We notice that  comparable performance is achieved without applying pooling operations for \textit{key} and \textit{query}. %This is likely due to the fact that at deeper stages of the ELGCA-Net having a hierarchical structure, it receives input features from the previous stage that are progressively attended to using PT attention.
However, we notice from Tab.~\ref{tbl:ablation_of_pooling} that the optimal combination of average and maximum pooling operations provides the best IoU score as 83.83\% where we use average pooling operation for \textit{query} feature maps and maximum pooling operation for \textit{key} feature maps.
% Other combinations of average and max pooling are not effective. 
%Whereas, when no pooling operation is applied to \textit{query} and \textit{key} features, 
%FLOPs are increased slightly, and 
%the IoU score is reduced to 82.80\%.
%We also perform an empirical study using different kernel sizes of the average pooling layer and observed that the kernel size of 3 and stride of 2 provide the optimal results.

\subsubsection{ {Ablation of losses applied during training of ELGCA}}
 {Finally, we perform an ablation study to show the impact of different losses utilized during the training of our proposed model.  From Tab. \ref{tbl:comp_of_losses}, it is notable that cross-entropy (CE) loss is more suitable to train our model compared to other losses.}

\section{Conclusion}
In this work, we present an efficient global and local context aggregation network (ELGC-Net) for remote sensing change detection problem. ELGC-Net utilizes the local and global contextual representations in an efficient way to capture enhanced semantic information while reducing the number of trainable parameters.
%and FLOPs.
Additionally, ELGC-Net does not require the use of a pre-trained backbone feature extractor. Extensive experimental results show the effectiveness of our approach. The proposed ELGC-Net outperforms on three challenging datasets and achieves state-of-the-art results. In addition, our lighter variant provides comparable performance while having reduced parameters and FLOPs.
The prospective future direction is to further increase the efficiency of the network and adapt the model for real-time change detection task on edge devices.

%\section*{Acknowledgment}
%The authors would like to thank and express their deepest gratitude to Mohamed bin Zayed University of Artificial Intelligence for the continuous support throughout the research journey.

\ifCLASSOPTIONcaptionsoff
  \newpage
\fi

\bibliographystyle{IEEEtran}
\bibliography{IEEEabrv, arxiv}

\end{document}